\documentclass[sigconf]{acmart}

\usepackage{tabularx}
\usepackage{algpseudocode}
\usepackage{algorithm}
\usepackage{CJKutf8}
\usepackage{balance}
\usepackage{adjustbox}
\algblock{Input}{EndInput}
\algnotext{EndInput}
\algblock{Output}{EndOutput}
\algnotext{EndOutput}

\AtBeginDocument{%
  \providecommand\BibTeX{{%
    \normalfont B\kern-0.5em{\scshape i\kern-0.25em b}\kern-0.8em\TeX}}}

\setcopyright{acmcopyright}
\copyrightyear{2018}
\acmYear{2018}
\acmDOI{XXXXXXX.XXXXXXX}

\acmConference[ACM ICMR 2022]{ACM International Conference on Multimedia Retrieval}{June 27--30, 2022}{Newark, NJ}
\acmPrice{15.00}
\acmISBN{978-1-4503-XXXX-X/18/06}

\copyrightyear{2022}
\acmYear{2022}
\setcopyright{acmcopyright}\acmConference[MAD '22]{Proceedings of the 1st International Workshop on Multimedia AI against Disinformation}{June 27--30, 2022}{Newark, NJ, USA}
\acmBooktitle{Proceedings of the 1st International Workshop on Multimedia AI against Disinformation (MAD '22), June 27--30, 2022, Newark, NJ, USA}
\acmPrice{15.00}
\acmDOI{10.1145/3512732.3533582}
\acmISBN{978-1-4503-9242-6/22/06}

\begin{document}
\fancyhead{}

\title{Cross-Forgery Analysis of Vision Transformers and CNNs \\ for Deepfake Image Detection}

%

\author{Davide Alessandro Coccomini}
\email{davidealessandro.coccomini@isti.cnr.it}
\orcid{XXXX}
\affiliation{%
  \institution{Inst. of Information Science and Technologies ``A. Faedo'' }
  \institution{Consiglio Nazionale delle Ricerche}
  \city{Pisa}
  \country{Italy}
}

\author{Roberto Caldelli}
\email{roberto.caldelli@unifi.it}
\orcid{XXXX}
\authornotemark[1]
\affiliation{%
  \institution{National Inter-University Consortium for Telecommunications (CNIT)}
  \city{Florence}
  \country{Italy}
\additionalaffiliation{
\institution{Universitas Mercatorum, Rome, Italy}
  }
}

\author{Fabrizio Falchi}
\email{fabrizio.falchi@isti.cnr.it}
\orcid{XXXX}
\affiliation{%
  \institution{Inst. of Information Science and Technologies ``A. Faedo'' }
  \institution{Consiglio Nazionale delle Ricerche}
  \city{Pisa}
  \country{Italy}
}

\author{Claudio Gennaro}
\email{claudio.gennaro@isti.cnr.it}
\orcid{XXXX}
\affiliation{%
  \institution{Inst. of Information Science and Technologies ``A. Faedo'' }
  \institution{Consiglio Nazionale delle Ricerche}
  \city{Pisa}
  \country{Italy}
}

\author{Giuseppe Amato}
\email{giuseppe.amato@isti.cnr.it}
\orcid{XXXX}
\affiliation{%
  \institution{Inst. of Information Science and Technologies ``A. Faedo'' }
  \institution{Consiglio Nazionale delle Ricerche}
  \city{Pisa}
  \country{Italy}
}


\begin{abstract}
Deepfake Generation Techniques are evolving at a rapid pace, making it possible to create realistic manipulated images and videos and endangering the serenity of modern society. The continual emergence of new and varied techniques brings with it a further problem to be faced, namely the ability of deepfake detection models to update themselves promptly in order to be able to identify manipulations carried out using even the most recent methods. This is an extremely complex problem to solve, as training a model requires large amounts of data, which are difficult to obtain if the deepfake generation method is too recent. Moreover, continuously retraining a network would be unfeasible. In this paper, we ask ourselves if, among the various deep learning techniques, there is one that is able to generalise the concept of deepfake to such an extent that it does not remain tied to one or more specific deepfake generation methods used in the training set. We compared a Vision Transformer with an EfficientNetV2 on a cross-forgery context based on the ForgeryNet dataset. From our experiments, It emerges that EfficientNetV2 has a greater tendency to specialize often obtaining better results on training methods while Vision Transformers exhibit a superior generalization ability that makes them more competent even on images generated with new methodologies. 

\end{abstract}



\maketitle

\section{Introduction}
The advancement of modern Deep Learning techniques is allowing society to evolve in many ways, helping in the achievement of increasingly stunning results in virtually every field. However, this progress also hides pitfalls with possible uses of Deep Learning that can be detrimental to the well-being of people. One of the most worrying emerging phenomena is undoubtedly that of deep fakes. These are images or videos manipulated by means of advanced Deep Learning techniques, to make the subjects filmed say or do things that they would never have said or done. The result of these techniques can then be used to destroy someone's reputation or to provoke conflict or manipulate the reality of events to one's advantage.

Distinguishing an image or video manipulated by these techniques from a genuine one has therefore become the goal of many researchers who have developed innovative methodologies, often also based on deep learning, to carry out what is called deepfake detection. In general, these techniques try to find any artefacts or anomalies that may be introduced during the manipulation process.

Machine Learning algorithms, however, need data to be trained, often in large quantities, and a number of datasets have sprung up trying to be as complete as possible in representing the various results that can be obtained with deepfake generation systems. In fact, there are numerous and varied techniques to manipulate multimedia content and ideally we would like to obtain a deepfake detector capable of identifying them regardless of the technique used for manipulation. Even more so, it would be ideal to have a system capable of identifying deepfakes generated by novel methods, examples of which are not present in the training dataset. In other words, we would like a deepfake detector capable of learning the general concept of deepfake and not simply being trained to recognise specific anomalies introduced by one or more specific deepfake generation methodologies. 
In this research, we therefore attempted to find out which of the main Deep Learning techniques was most capable of generalising the concept of deepfake and therefore proved robust in identifying images manipulated by methods it had never been trained on. The comparison was made between the two categories of neural networks used in this field, Vision Transformer and Convolutional Neural Networks.  Our experiments showed that the former were less inclined to specialise in a specific method and were able to obtain consistent results even on deepfakes generated with novel methodologies. On the other hand, the Convolutional Neural Networks appear able to reach better performances in terms of accuracy on the training methods.

\section{Related Works}
\subsection{Deepfake Generation}
Deepfake Generation techniques are the set of methods used to manipulate a human face in order to make it appear different or to replace its identity in a realistic manner. There are two main categories of approaches, those based on Variational AutoEncoders (VAEs) \cite{kingma2014autoencoding} and those based on Generative Adversarial Networks (GANs) \cite{goodfellow2014generative}.
Methods based on VAEs use encoder-decoder pairs to decompose and recompose two distinct faces. By then swapping the decoders, it is possible to obtain one of the two faces from the other face, yielding quite credible results.

The GAN-based methods use two different networks. The first one, network called discriminator, is trained to classify if an image is fake or not, and a second network called generator that instead must succeed, starting from a noisy image, to generate one sufficiently credible to deceive its counterpart. GANs have been particularly effective in the field of deepfake generation, with excellent results achieved with methodologies based on networks such as DiscoGAN \cite{kim2017learning}, StarGAN \cite{8579014} and StyleGAN-V2 \cite{karras2020analyzing}.

Regardless of the technique used to carry out the manipulation, the various deepfake generation approaches are distinguished according to the specific way in which the image is modified. Among these the following stand out:

\begin{itemize}
    \item \textit{Face Transfer}: it transfers both identity-aware and identity-agnostic content (e.g. expression and pose) from a source face to the target face;
    \item \textit{Face Swap}: it transfers the identity of the source face to the target face while preserving identity-agnostic content;
    \item \textit{Face Stacked Manipulation (FSM)}: set of methodologies some that transfer both the identity and the attributes of the target on the source while others that alter the attributes of the swapped target after the transfer of the identity;
    \item \textit{Face Reenactment}: it preserves the identity of the source subject but manipulates the intrinsic attributes such as mouth or expression;
    \item \textit{Face Editing}: it edits external attributes such as age, gender or ethnicity.
\end{itemize}

\subsection{Deepfake Detection}
With this wide variety of deepfake generation methodologies and their increasing effectiveness, it has therefore become extremely important to develop systems to distinguish a manipulated image from a real one. This is a problem that also affects other areas such as text, with recent work such as \cite{fagni2021tweepfake} analyzing deepfakes in tweets to identify fake content in social networks.
Anyway, as these techniques are often applied on videos, many video deepfake detectors emerged. Some recent works proposed the exploitation of temporal information to recognise inconsistencies. For example \cite{CALDELLI202131} try to catch motion dissimilarities in the temporal structure of a video sequence by exploiting optical flow fields. However the majority of approaches are frame-based, classifying the video frame by frame.
In order to train effective deep learning models for deepfake detection, a number of datasets have been created over the years, including the first DF-TIMIT \cite{korshunov2018deepfakes}, UADFC \cite{yang2019exposing} and FaceForensics++ \cite{rossler2019faceforensics}, Celeb-DF \cite{li2020celebdf}, Google Deepfake Detection Dataset \cite{googledf} and the more recent DFDC \cite{dolhansky2020deepfake}, Deepforensics \cite{jiang2020deeperforensics10} and ForgeryNet \cite{forgerynet}. The latter dataset is the most complete, largest and includes the greater variety of existing deepfake generation methods, since it is still recently published there are not many papers based on it.
Much research has been carried out exploiting the DFDC dataset as it is one of the most complete and challenging in circulation and a specific type of convolutional neural network has emerged as particularly effective in fulfilling the task, the EfficientNet. The latter is the basis of many solutions that have obtained state-of-the-art results on the cited dataset such as the winning solution of the deepfake detection challenge \cite{dfdc_solution}. With the advent of Vision Transformers and their successes in the field of Computer Vision, some interesting deepfake detection solutions have emerged. For example, the method presented in \cite{wodajo2021deepfake} which obtained good results by combining Transformers with convolutional networks used to extract patches from faces. Also interesting is the work done to exploit a pretrained EfficientNet B7 with a Vision Transformers by training it through distillation presented in \cite{heo2021deepfake}.
A recent work on merging different types of Vision Transformers such as the Cross Vision Transformer \cite{chen2021crossvit} and EfficientNet B0 is presented in \cite{coccomini2022combining}.
EfficientNet has recently been further improved with the introduction of EfficientNetV2 \cite{tan2021efficientnetv2}, a version of EfficientNet that is more optimised for smaller models, faster training and better ImageNet \cite{5206848} accuracy than its predecessor and some Vision Transformers.

\begin{figure}[t]
  \centering
  \includegraphics[width=0.4\textwidth]{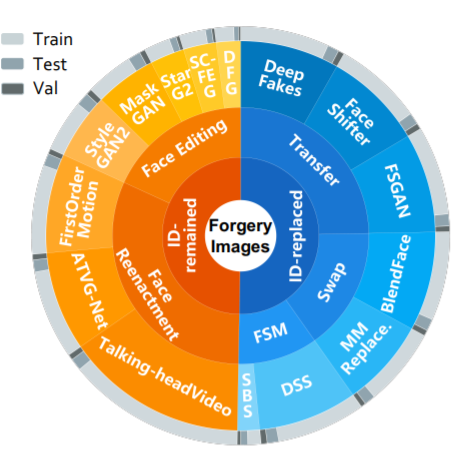}
  \caption{Visualization of the various Deepfake generation techniques within the dataset grouped by category \cite{forgerynet}}.
  \label{fig:forgery_methods}
\end{figure}

\begin{figure*}[hbt!]
  \centering
  \includegraphics[width=\textwidth]{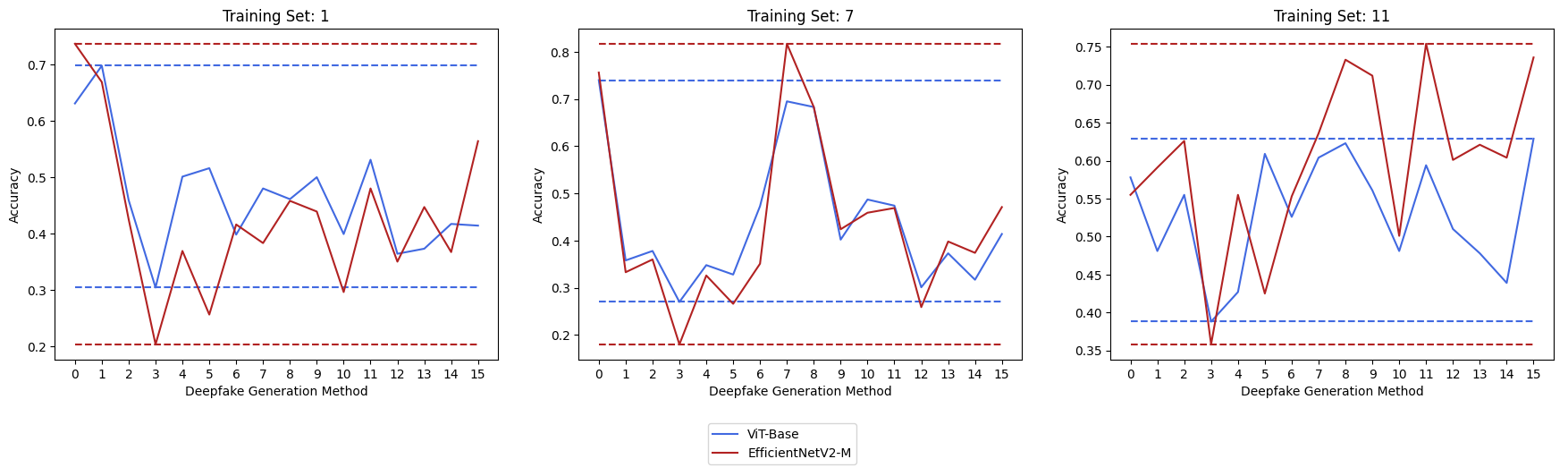}
  \caption{Line plots representing the accuracy values obtained by ViT-Base (blue) and EfficientnetV2-M (red) on the test set, trained on training sets consisting of real images and from left to right on images generated with the FaceShifter, Talking Head Video and StyleGAN2 methodology respectively.
Observing the space between horizontal lines it is possible to notice how the variance of the Vision Transformer is lower than that of the CNN counterpart obtaining closer accuracies on different and also unseen methods.  On the horizontal axis the 0 represent the real images and number from 1-15 the images generated with different generation methods.}
  \label{fig:single_methods}
\end{figure*}

\section{Approach}
In this section, we analyze in detail the dataset and the models used to carry out our experiments.

\subsection{Dataset}
To be able to validate the ability of a neural network to detect deepfakes generated by methods other than those used for the construction of the training set, it is necessary to use a dataset containing a multitude of deepfake generation methods and keeps track of them.
For this reason the dataset selected to carry out the experiments is ForgeryNet \cite{forgerynet}, one of the widest deepfake datasets available. 
ForgeryNet consists of 2.9M images and 220k video clips. For our experiments we will only use the set of images for which we have, for each of them, an associated label identifying if it has been manipulated or not and the method adopted to perform such manipulation. The fake images are generated through the use of 15 different manipulation approaches \cite{8953690,9156396,392910,9010058,9156570,9157722,9156865,9010341,Siarohin2019FirstOM, Petrov2020DeepFaceLabAS} with more than 36 mix-perturbations on more than 4300 distinct subjects. Examples of applied perturbations are optical distortion, multiplicative noise, radom compression, blur and many others shown in more detail in the ForgeryNet paper. This variety therefore allows for the most comprehensive comparison possible between the two categories of neural networks. The methodologies used can be grouped into two macro-categories, \textit{ID-remained} and \textit{ID-replaced}, as shown in Figure \ref{fig:forgery_methods}. In the first case the identity of the subject in the image is not replaced but only manipulations are carried out on his face. In the second category, on the other hand, the identity is replaced by transferring a different face from the one actually present in the image. In turn, these categories are divided into 5 sub-categories: Face-Reenactment and Face Editing, belonging to the ID-remained category and Face Transfer, Face Swap and FSM, belonging instead to the ID-replaced category. All these approaches represent a large part of the main methods of deepfake generation known to date.
The images in ForgeryNet also include people in a variety of contexts. To make the task of the two networks simpler, as in many other deepfake detection methods, a face extraction phase is carried out through the use of a state-of-the-art face detector, MTCNN \cite{zhang2016joint}. The considered models are trained and tested on a face basis and we performed data augmentation like in \cite{dfdc_solution}. Differently from them, we extracted the faces so that they were always squared and without padding. We used the Albumentations library \cite{2018arXiv180906839B}, and during the training, we randomly applied common transformations. In particular, every time an image is passed to the network in training phase, this is resized randomly with three types of isotropic resize that differ for type of interpolation used (area, cubic or linear). After that, transformations are applied randomly, namely: image compression, gaussian noise, horizontal flip, brightness or saturation distortion, grayscale conversion and shift, rotation or scaling.

\subsection{Considered Architectures}
In this research, we want to compare the cross-forgery generalisation capability of Convolutional Neural Networks, a category of neural networks widely used in this and many other Computer Vision tasks, with that of Vision Transformers \cite{50650}, a more recent type of deep learning model that is proving to be particularly competitive.
For the first category, an EfficientNetV2-M \cite{https://doi.org/10.48550/arxiv.2104.00298} was selected, a new version of the well-known EfficientNet that is more powerful and lighter. EfficientNets are widely used in deepfake detection and still form the basis of many state-of-the-art methods on the industry's leading datasets.
The counterpart used is instead a ViT-Base, one of the first Vision Transformers presented and of similar dimensions with respect to the Convolutional network considered. 
Both networks are pretrained on ImageNet-21k and have been fine-tuned on sub-datasets extracted from ForgeryNet. The sub-datasets were constructed maintaining an almost perfect balance between fake and real images. In addition, only faces detected with a confidence level higher than 95\% were considered in order to reduce the risk of false detection. The networks were trained freezing the weights of all blocks except for the last two which are specialised on the downstream task.

\section{Experiments}
In this section, we describe all the experiments carried out in our research. 
The experiments are subdivided in two parts, the first in which we will use genuine images and images generated with a single method of deepfake generation at a time, and the second part in which instead we will consider more methods of deepfake generation, belonging to the same category, in the training phase.
Since the labels of the ForgeryNet test set were not yet released at the time of the experiments, the validation set of this dataset, from now on called test set, was used to carry out all the tests while in the training phase, a portion equal to 10\%, always the same for all the models, was selected from the sub-dataset considered. The latter will be called validation set from now on. The models were trained for a maximum of 50 epochs and a patience of 5 epochs on the validation set, using an SGD optimiser with a learning rate of 0.1 decreasing with a step size of 15 and a gamma of 0.1.

\subsection{Single Method Training}

\begin{table}
  \includegraphics[width=0.3\textwidth]{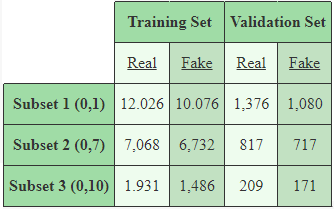}
  \caption{Number of images per subset in the Single Methods experiment.}
  \label{tab:datasets1}
\end{table}

In this section, we describe the process used to investigate the ability of a model, trained on real images and images manipulated with a single deepfake generation method, to generalise the concept of deepfake to the point of recognizing images tampered with by other methods.
\begin{figure}[h]
    \includegraphics[width=0.45\textwidth]{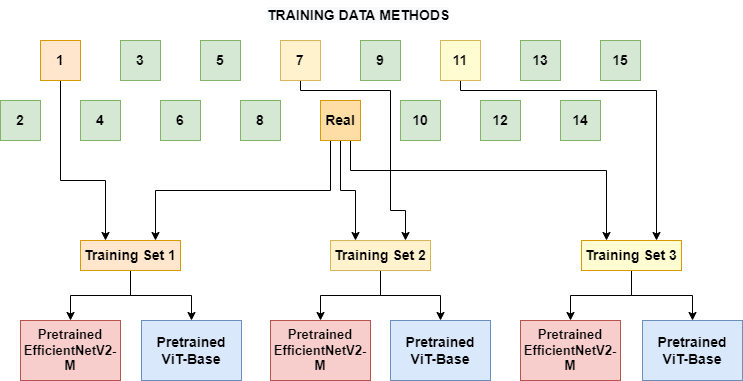}
    \caption{Training set construction for the Single-Method approach.}
  \label{fig:single_method_dataset}
\end{figure}\\
To perform this first comparison the two models were fine tuned on three subdatasets as shown in Figure \ref{fig:single_method_dataset}. All of them contained unmanipulated images but also tampered images with a specific technique for each subdataset. The three techniques used are FaceShifter (1), Talking Head Video (7) and StyleGAN2 (11). These were selected as being quite different from each other so as to validate the effectiveness of the two networks on more varied manipulation approaches. As shown in Table \ref{tab:datasets1}, the sizes of the three datasets are quite varied but always well balanced between the two classes. In this experiment then, the models will only see anomalies introduced by one specific deepfakes generation method at a time.
The models trained with the three sub-datasets were then tested with the images in the test set, but also considering the generation methods not used in the training set.

As can be seen in the three plots in Figure \ref{fig:single_methods} the Vision Transformers turn out to be more stable and less specialized. This model, even if reaches better results on the training methods, tends to have values of accuracy closer between the various methods of deepfake generation. On the other hand the EfficientNet often obtains higher accuracy than the Vision Transformer on the training methods but achieves much poorer accuracies on the others. For example, we can see a marked advantage of the EfficientNet over both method 7 and method 11 in the respective charts.

\subsection{Multiple Methods Training}

\begin{figure}[h]
    \includegraphics[width=0.45\textwidth]{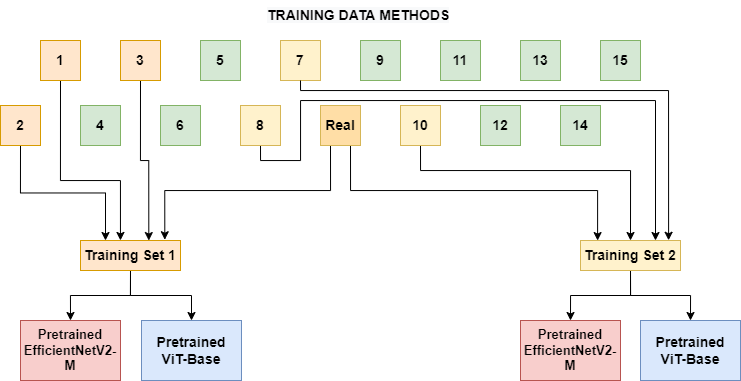}
    \caption{Training set construction for the Multiple-Methods approach.}
  \label{fig:multiple_methods_dataset}
\end{figure}

A further experiment was carried out by training on real images and on images manipulated with a group of methods belonging to the same category as shown in Figure \ref{fig:multiple_methods_dataset}. This choice derives from the fact that one of the two networks may be able to generalise even better in the presence of different generation methods, which therefore hopefully introduce a greater variety of artefacts. 

\begin{figure*}[h]
  \centering
  \includegraphics[width=0.9\textwidth]{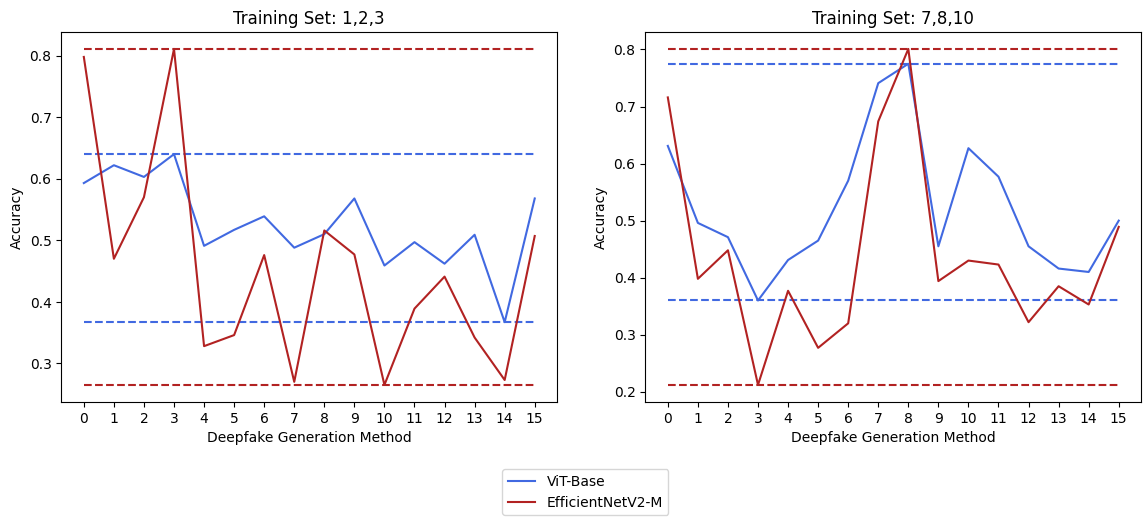}
  \caption{Line plots representing the accuracy values obtained by ViT-Base and EfficientNetV2-M on the test set, trained on a training set of real images and on images manipulated with methods belonging to the Transfer category (left) and those belonging to the Face Reenactment category (right). On the horizontal axis the 0 represent the real images and number from 1-15 the images generated with different generation methods.}
  \label{fig:multiple_methods}
\end{figure*}

\begin{table}
  \includegraphics[width=0.33\textwidth]{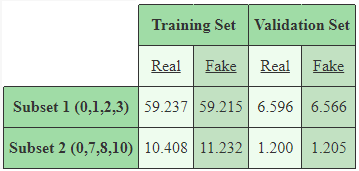}
  \caption{Number of images per subset in the Multiple Methods experiment.}
  \label{tab:datasets2}
\end{table}

For the first experiment, the training methods considered were those belonging to the Transfer category, i.e. FaceShifter (1), FS-GAN (2) and Deepfakes (3), as well as unmanipulated images. For the second experiment the methods belonging to the Face Reenactment category were used, namely Talking Head Video (7), ATVG-net (8) and First Order Motion (10). As shown in Table \ref{tab:datasets2}, both subsets are well balanced but differ in size. The former is in fact considerably larger, with about 120,000 traning images and representing the largest of the sets considered in the experiments, while the latter is smaller in size.

\begin{table*}
  \centering
  \includegraphics[width=\textwidth]{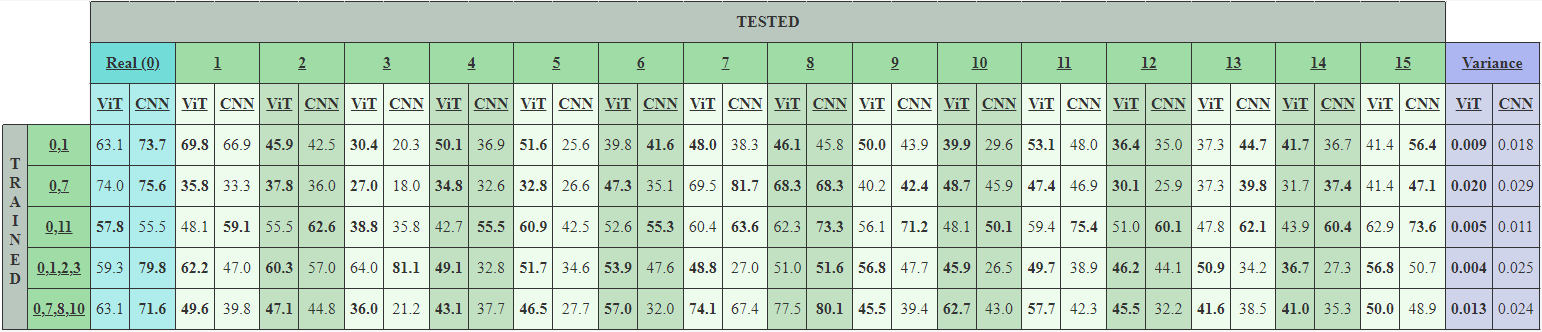}
  \caption{Table summarizing the accuracies obtained by the models on real test images (column 0) and on those manipulated with all deepfake generation methods considered (columns 1-15). The last column of the table contains the calculated variance values between the accuracies obtained by the models on the test set in the various deepfakes generation methods for each training sub-dataset.}
  \label{tab:summary}
\end{table*}

Again, ViT-Base is found to have a significantly lower variance than EfficientNetV2-M in both experiments conducted, as can be seen from the horizontal lines in the charts in Figure \ref{fig:multiple_methods}. There is also a tendency for EfficientNet to focus on a subset of the methods presented in the training set. For example, in the case of the dataset consisting of images manipulated with Transfer techniques, the convolutional network obtains an accuracy value of 81.1\% on method 3, while it remains particularly low on methods 1 and 2, with accuracy values of 47.0\% and 57.0\% respectively. On the other hand, the Vision Tranformer obtains rather similar accuracy values on all three methods considered, although they are lower and are always around 62.0\%.

The same behaviour can be observed in the second experiment. In this case, the EfficientNet reaches an accuracy of 80.1\% on method 8 but drops drastically to 67.4\% and 43.0\% on the other two training methods, respectively 7 and 10. The Vision Transformer instead once again proves to be more stable with an accuracy of 74.1\% and 77.5\% on methods 7 and 8 and a less marked drop in performance on method 10 reaching 62.7\%. In simple terms, in all plots the blue line representing Vision Transformers tends to remain higher than the red line representing EfficientNets.

To conclude, although the accuracies therefore tend to be rather low on the various novel methods, there is a tendency for EfficientNets to perform better on training methods than Vision Transformers, often achieving higher accuracies, but to generalise worse on novel methods.

\subsection{Final Results}

To numerically evaluate which of the two models is less likely to specialize in the deepfake generation methods used to build the training set, we chose to calculate the variance between the accuracies on the test set.
More in detail, considering the list of $16$ accuracies, one for each deepfake generation method plus the one obtained on the unmanipulated images, obtained from each network on the test images, the variance $\sigma^2$ was calculated as follows:

$$\sigma^2 = \frac{\sum_{i=1}^{n} (x_i - \mu)^2}{n} $$

where $n=16$ is the number of accuracies, $x_i$ are the accuracy values and $\mu$ is their mean.

A model with a lower variance will have obtained more similar accuracies between all deepfake generation methods regardless of whether they have been included in the training set and therefore will have learned the concept of deepfake more generally without specializing too much on the specific anomalies introduced by a method. 
From the data reported in the table \ref{tab:summary} it can be seen that regardless of the methods used to build the training set, the variance associated with the Vision Transformers is always lower. On the other hand, in almost all cases the EfficientNet achieves higher levels of accuracy on training methods probably because it has learned to better recognize the specific anomalies introduced in these methods and all images containing different anomalies are considered non-deepfake.
The case of the training on method 11 only, StyleGAN2, is interesting. In this case, both models have a marked decrease in performance on unmanipulated images. This probably derives from the fact that this specific method is particularly effective and introduces fewer anomalies than others present in the dataset, thus making the difference between a real image and a manipulated one more nuanced.

\balance

\section{Conclusions}
In this paper, we conducted a cross-forgery analysis to identify the most suitable deep learning architecture to tackle the deepfake detection task. The experiments carried out allowed us to have a first confirmation of the tendency of the Vision Transformers to better generalise the concept of deepfake, exhibiting less bias towards specific anomalies introduced by one or more deepfake generation techniques and thus making them more suitable to be applied in a real-world context. On the other hand, the convolutional networks and in particular, the EfficientNet, seem to be more prone to specialization, making them more applicable in contexts in which one wants to carry out deepfake detection, excluding the possibility that images manipulated with unpublished techniques can be introduced.
Investigating the different ways of approaching the problem of the various deepfake detection solutions, not limiting ourselves exclusively to evaluating accuracy metrics on a subset of well-known and studied methods, is therefore fundamental to creating robust and long-lasting systems. With this research we have taken a step forward, highlighting in greater detail the behaviour of the main architectures used in the sector.


\begin{acks}
This work was partially supported by the AI4Media project, funded by the EC (H2020 - Contract n. 951911).
\end{acks}

\bibliographystyle{ACM-Reference-Format}
\bibliography{main}


\begin{thebibliography}{36}


\ifx \showCODEN    \undefined \def \showCODEN     #1{\unskip}     \fi
\ifx \showDOI      \undefined \def \showDOI       #1{#1}\fi
\ifx \showISBNx    \undefined \def \showISBNx     #1{\unskip}     \fi
\ifx \showISBNxiii \undefined \def \showISBNxiii  #1{\unskip}     \fi
\ifx \showISSN     \undefined \def \showISSN      #1{\unskip}     \fi
\ifx \showLCCN     \undefined \def \showLCCN      #1{\unskip}     \fi
\ifx \shownote     \undefined \def \shownote      #1{#1}          \fi
\ifx \showarticletitle \undefined \def \showarticletitle #1{#1}   \fi
\ifx \showURL      \undefined \def \showURL       {\relax}        \fi
\providecommand\bibfield[2]{#2}
\providecommand\bibinfo[2]{#2}
\providecommand\natexlab[1]{#1}
\providecommand\showeprint[2][]{arXiv:#2}

\bibitem[Buslaev et~al\mbox{.}(2018)]%
        {2018arXiv180906839B}
\bibfield{author}{\bibinfo{person}{A. Buslaev}, \bibinfo{person}{A. Parinov},
  \bibinfo{person}{E. Khvedchenya}, \bibinfo{person}{V.~I. Iglovikov}, {and}
  \bibinfo{person}{A.~A. Kalinin}.} \bibinfo{year}{2018}\natexlab{}.
\newblock \showarticletitle{{Albumentations: fast and flexible image
  augmentations}}.
\newblock \bibinfo{journal}{\emph{ArXiv e-prints}} (\bibinfo{year}{2018}).
\newblock
\showeprint{1809.06839}


\bibitem[Caldelli et~al\mbox{.}(2021)]%
        {CALDELLI202131}
\bibfield{author}{\bibinfo{person}{Roberto Caldelli}, \bibinfo{person}{Leonardo
  Galteri}, \bibinfo{person}{Irene Amerini}, {and} \bibinfo{person}{Alberto
  {Del Bimbo}}.} \bibinfo{year}{2021}\natexlab{}.
\newblock \showarticletitle{Optical Flow based CNN for detection of unlearnt
  deepfake manipulations}.
\newblock \bibinfo{journal}{\emph{Pattern Recognition Letters}}
  \bibinfo{volume}{146} (\bibinfo{year}{2021}), \bibinfo{pages}{31--37}.
\newblock
\showISSN{0167-8655}
\urldef\tempurl%
\url{https://doi.org/10.1016/j.patrec.2021.03.005}
\showDOI{\tempurl}


\bibitem[Chen et~al\mbox{.}(2021)]%
        {chen2021crossvit}
\bibfield{author}{\bibinfo{person}{Chun-Fu Chen}, \bibinfo{person}{Quanfu Fan},
  {and} \bibinfo{person}{Rameswar Panda}.} \bibinfo{year}{2021}\natexlab{}.
\newblock \showarticletitle{Crossvit: Cross-attention multi-scale vision
  transformer for image classification}.
\newblock \bibinfo{journal}{\emph{arXiv preprint arXiv:2103.14899}}
  (\bibinfo{year}{2021}).
\newblock


\bibitem[Chen et~al\mbox{.}(2019)]%
        {8953690}
\bibfield{author}{\bibinfo{person}{Lele Chen}, \bibinfo{person}{Ross~K.
  Maddox}, \bibinfo{person}{Zhiyao Duan}, {and} \bibinfo{person}{Chenliang
  Xu}.} \bibinfo{year}{2019}\natexlab{}.
\newblock \showarticletitle{Hierarchical Cross-Modal Talking Face Generation
  With Dynamic Pixel-Wise Loss}. In \bibinfo{booktitle}{\emph{2019 IEEE/CVF
  Conference on Computer Vision and Pattern Recognition (CVPR)}}.
  \bibinfo{pages}{7824--7833}.
\newblock
\urldef\tempurl%
\url{https://doi.org/10.1109/CVPR.2019.00802}
\showDOI{\tempurl}


\bibitem[Choi et~al\mbox{.}(2018)]%
        {8579014}
\bibfield{author}{\bibinfo{person}{Yunjey Choi}, \bibinfo{person}{Minje Choi},
  \bibinfo{person}{Munyoung Kim}, \bibinfo{person}{Jung-Woo Ha},
  \bibinfo{person}{Sunghun Kim}, {and} \bibinfo{person}{Jaegul Choo}.}
  \bibinfo{year}{2018}\natexlab{}.
\newblock \showarticletitle{StarGAN: Unified Generative Adversarial Networks
  for Multi-domain Image-to-Image Translation}. In
  \bibinfo{booktitle}{\emph{2018 IEEE/CVF Conference on Computer Vision and
  Pattern Recognition}}. \bibinfo{pages}{8789--8797}.
\newblock
\urldef\tempurl%
\url{https://doi.org/10.1109/CVPR.2018.00916}
\showDOI{\tempurl}


\bibitem[Coccomini et~al\mbox{.}(2022)]%
        {coccomini2022combining}
\bibfield{author}{\bibinfo{person}{Davide Coccomini}, \bibinfo{person}{Nicola
  Messina}, \bibinfo{person}{Claudio Gennaro}, {and} \bibinfo{person}{Fabrizio
  Falchi}.} \bibinfo{year}{2022}\natexlab{}.
\newblock \bibinfo{title}{Combining EfficientNet and Vision Transformers for
  Video Deepfake Detection}.
\newblock
\newblock
\showeprint[arxiv]{2107.02612}~[cs.CV]


\bibitem[Deng et~al\mbox{.}(2009)]%
        {5206848}
\bibfield{author}{\bibinfo{person}{Jia Deng}, \bibinfo{person}{Wei Dong},
  \bibinfo{person}{Richard Socher}, \bibinfo{person}{Li-Jia Li},
  \bibinfo{person}{Kai Li}, {and} \bibinfo{person}{Li Fei-Fei}.}
  \bibinfo{year}{2009}\natexlab{}.
\newblock \showarticletitle{ImageNet: A large-scale hierarchical image
  database}. In \bibinfo{booktitle}{\emph{2009 IEEE Conference on Computer
  Vision and Pattern Recognition}}. \bibinfo{pages}{248--255}.
\newblock
\urldef\tempurl%
\url{https://doi.org/10.1109/CVPR.2009.5206848}
\showDOI{\tempurl}


\bibitem[Deng et~al\mbox{.}(2020)]%
        {9156396}
\bibfield{author}{\bibinfo{person}{Yu Deng}, \bibinfo{person}{Jiaolong Yang},
  \bibinfo{person}{Dong Chen}, \bibinfo{person}{Fang Wen}, {and}
  \bibinfo{person}{Xin Tong}.} \bibinfo{year}{2020}\natexlab{}.
\newblock \showarticletitle{Disentangled and Controllable Face Image Generation
  via 3D Imitative-Contrastive Learning}. In \bibinfo{booktitle}{\emph{2020
  IEEE/CVF Conference on Computer Vision and Pattern Recognition (CVPR)}}.
  \bibinfo{pages}{5153--5162}.
\newblock
\urldef\tempurl%
\url{https://doi.org/10.1109/CVPR42600.2020.00520}
\showDOI{\tempurl}


\bibitem[Dolhansky et~al\mbox{.}(2020)]%
        {dolhansky2020deepfake}
\bibfield{author}{\bibinfo{person}{Brian Dolhansky}, \bibinfo{person}{Joanna
  Bitton}, \bibinfo{person}{Ben Pflaum}, \bibinfo{person}{Jikuo Lu},
  \bibinfo{person}{Russ Howes}, \bibinfo{person}{Menglin Wang}, {and}
  \bibinfo{person}{Cristian~Canton Ferrer}.} \bibinfo{year}{2020}\natexlab{}.
\newblock \showarticletitle{The deepfake detection challenge (dfdc) dataset}.
\newblock \bibinfo{journal}{\emph{arXiv preprint arXiv:2006.07397}}
  (\bibinfo{year}{2020}).
\newblock


\bibitem[Dufour and Gully(2019)]%
        {googledf}
\bibfield{author}{\bibinfo{person}{Nick Dufour} {and} \bibinfo{person}{Andrew
  Gully}.} \bibinfo{year}{2019}\natexlab{}.
\newblock \bibinfo{title}{Contributing data to deep-fake detection research}.
\newblock
\newblock
\urldef\tempurl%
\url{https://ai.googleblog.com/2019/09/contributing-data-to-deepfake-detection.html}
\showURL{%
\tempurl}


\bibitem[Fagni et~al\mbox{.}(2021)]%
        {fagni2021tweepfake}
\bibfield{author}{\bibinfo{person}{Tiziano Fagni}, \bibinfo{person}{Fabrizio
  Falchi}, \bibinfo{person}{Margherita Gambini}, \bibinfo{person}{Antonio
  Martella}, {and} \bibinfo{person}{Maurizio Tesconi}.}
  \bibinfo{year}{2021}\natexlab{}.
\newblock \showarticletitle{TweepFake: About detecting deepfake tweets}.
\newblock \bibinfo{journal}{\emph{Plos one}} \bibinfo{volume}{16},
  \bibinfo{number}{5} (\bibinfo{year}{2021}), \bibinfo{pages}{e0251415}.
\newblock


\bibitem[Fried et~al\mbox{.}(2019)]%
        {392910}
\bibfield{author}{\bibinfo{person}{Ohad Fried}, \bibinfo{person}{Ayush Tewari},
  \bibinfo{person}{Michael Zollhöfer}, \bibinfo{person}{Adam Finkelstein},
  \bibinfo{person}{Eli Shechtman}, \bibinfo{person}{Dan Goldman},
  \bibinfo{person}{Kyle Genova}, \bibinfo{person}{Zeyu Jin},
  \bibinfo{person}{Christian Theobalt}, {and} \bibinfo{person}{Maneesh
  Agrawala}.} \bibinfo{year}{2019}\natexlab{}.
\newblock \bibinfo{title}{Text-based Editing of Talking-head Video}.
\newblock
\newblock


\bibitem[Goodfellow et~al\mbox{.}(2014)]%
        {goodfellow2014generative}
\bibfield{author}{\bibinfo{person}{Ian~J. Goodfellow}, \bibinfo{person}{Jean
  Pouget-Abadie}, \bibinfo{person}{Mehdi Mirza}, \bibinfo{person}{Bing Xu},
  \bibinfo{person}{David Warde-Farley}, \bibinfo{person}{Sherjil Ozair},
  \bibinfo{person}{Aaron Courville}, {and} \bibinfo{person}{Yoshua Bengio}.}
  \bibinfo{year}{2014}\natexlab{}.
\newblock \showarticletitle{Generative Adversarial Networks}. In
  \bibinfo{booktitle}{\emph{Advances in neural information processing systems
  27}}.
\newblock
\showeprint[arxiv]{1406.2661}~[stat.ML]


\bibitem[He et~al\mbox{.}(2021)]%
        {forgerynet}
\bibfield{author}{\bibinfo{person}{Yinan He}, \bibinfo{person}{Bei Gan},
  \bibinfo{person}{Siyu Chen}, \bibinfo{person}{Yichun Zhou},
  \bibinfo{person}{Guojun Yin}, \bibinfo{person}{Luchuan Song},
  \bibinfo{person}{Lu Sheng}, \bibinfo{person}{Jing Shao}, {and}
  \bibinfo{person}{Ziwei Liu}.} \bibinfo{year}{2021}\natexlab{}.
\newblock \showarticletitle{ForgeryNet: A Versatile Benchmark for Comprehensive
  Forgery Analysis}. In \bibinfo{booktitle}{\emph{2021 IEEE/CVF Conference on
  Computer Vision and Pattern Recognition (CVPR)}}.
  \bibinfo{pages}{4358--4367}.
\newblock
\urldef\tempurl%
\url{https://doi.org/10.1109/CVPR46437.2021.00434}
\showDOI{\tempurl}


\bibitem[Heo et~al\mbox{.}(2021)]%
        {heo2021deepfake}
\bibfield{author}{\bibinfo{person}{Young-Jin Heo}, \bibinfo{person}{Young-Ju
  Choi}, \bibinfo{person}{Young-Woon Lee}, {and} \bibinfo{person}{Byung-Gyu
  Kim}.} \bibinfo{year}{2021}\natexlab{}.
\newblock \showarticletitle{Deepfake Detection Scheme Based on Vision
  Transformer and Distillation}.
\newblock \bibinfo{journal}{\emph{arXiv preprint arXiv:2104.01353}}
  (\bibinfo{year}{2021}).
\newblock


\bibitem[Jiang et~al\mbox{.}(2020)]%
        {jiang2020deeperforensics10}
\bibfield{author}{\bibinfo{person}{Liming Jiang}, \bibinfo{person}{Ren Li},
  \bibinfo{person}{Wayne Wu}, \bibinfo{person}{Chen Qian}, {and}
  \bibinfo{person}{Chen~Change Loy}.} \bibinfo{year}{2020}\natexlab{}.
\newblock \showarticletitle{Deeperforensics-1.0: A large-scale dataset for
  real-world face forgery detection}. In \bibinfo{booktitle}{\emph{Proceedings
  of the IEEE/CVF Conference on Computer Vision and Pattern Recognition}}.
  \bibinfo{pages}{2889--2898}.
\newblock


\bibitem[Jo and Park(2019)]%
        {9010058}
\bibfield{author}{\bibinfo{person}{Youngjoo Jo} {and} \bibinfo{person}{Jongyoul
  Park}.} \bibinfo{year}{2019}\natexlab{}.
\newblock \showarticletitle{SC-FEGAN: Face Editing Generative Adversarial
  Network With User’s Sketch and Color}. In \bibinfo{booktitle}{\emph{2019
  IEEE/CVF International Conference on Computer Vision (ICCV)}}.
  \bibinfo{pages}{1745--1753}.
\newblock
\urldef\tempurl%
\url{https://doi.org/10.1109/ICCV.2019.00183}
\showDOI{\tempurl}


\bibitem[Karras et~al\mbox{.}(2020a)]%
        {karras2020analyzing}
\bibfield{author}{\bibinfo{person}{Tero Karras}, \bibinfo{person}{Samuli
  Laine}, \bibinfo{person}{Miika Aittala}, \bibinfo{person}{Janne Hellsten},
  \bibinfo{person}{Jaakko Lehtinen}, {and} \bibinfo{person}{Timo Aila}.}
  \bibinfo{year}{2020}\natexlab{a}.
\newblock \showarticletitle{Analyzing and improving the image quality of
  stylegan}. In \bibinfo{booktitle}{\emph{Proceedings of the IEEE/CVF
  Conference on Computer Vision and Pattern Recognition}}.
  \bibinfo{pages}{8110--8119}.
\newblock


\bibitem[Karras et~al\mbox{.}(2020b)]%
        {9156570}
\bibfield{author}{\bibinfo{person}{Tero Karras}, \bibinfo{person}{Samuli
  Laine}, \bibinfo{person}{Miika Aittala}, \bibinfo{person}{Janne Hellsten},
  \bibinfo{person}{Jaakko Lehtinen}, {and} \bibinfo{person}{Timo Aila}.}
  \bibinfo{year}{2020}\natexlab{b}.
\newblock \showarticletitle{Analyzing and Improving the Image Quality of
  StyleGAN}. In \bibinfo{booktitle}{\emph{2020 IEEE/CVF Conference on Computer
  Vision and Pattern Recognition (CVPR)}}. \bibinfo{pages}{8107--8116}.
\newblock
\urldef\tempurl%
\url{https://doi.org/10.1109/CVPR42600.2020.00813}
\showDOI{\tempurl}


\bibitem[Kim et~al\mbox{.}(2017)]%
        {kim2017learning}
\bibfield{author}{\bibinfo{person}{Taeksoo Kim}, \bibinfo{person}{Moonsu Cha},
  \bibinfo{person}{Hyunsoo Kim}, \bibinfo{person}{Jung~Kwon Lee}, {and}
  \bibinfo{person}{Jiwon Kim}.} \bibinfo{year}{2017}\natexlab{}.
\newblock \showarticletitle{Learning to discover cross-domain relations with
  generative adversarial networks}. In \bibinfo{booktitle}{\emph{International
  Conference on Machine Learning}}. PMLR, \bibinfo{pages}{1857--1865}.
\newblock


\bibitem[Kingma and Welling(2013)]%
        {kingma2014autoencoding}
\bibfield{author}{\bibinfo{person}{Diederik~P Kingma} {and}
  \bibinfo{person}{Max Welling}.} \bibinfo{year}{2013}\natexlab{}.
\newblock \showarticletitle{Auto-encoding variational bayes}.
\newblock \bibinfo{journal}{\emph{arXiv preprint arXiv:1312.6114}}
  (\bibinfo{year}{2013}).
\newblock


\bibitem[Kolesnikov et~al\mbox{.}(2021)]%
        {50650}
\bibfield{author}{\bibinfo{person}{Alexander Kolesnikov},
  \bibinfo{person}{Alexey Dosovitskiy}, \bibinfo{person}{Dirk Weissenborn},
  \bibinfo{person}{Georg Heigold}, \bibinfo{person}{Jakob Uszkoreit},
  \bibinfo{person}{Lucas Beyer}, \bibinfo{person}{Matthias Minderer},
  \bibinfo{person}{Mostafa Dehghani}, \bibinfo{person}{Neil Houlsby},
  \bibinfo{person}{Sylvain Gelly}, \bibinfo{person}{Thomas Unterthiner}, {and}
  \bibinfo{person}{Xiaohua Zhai}.} \bibinfo{year}{2021}\natexlab{}.
\newblock \showarticletitle{An Image is Worth 16x16 Words: Transformers for
  Image Recognition at Scale}.
\newblock


\bibitem[Korshunov and Marcel(2018)]%
        {korshunov2018deepfakes}
\bibfield{author}{\bibinfo{person}{Pavel Korshunov} {and}
  \bibinfo{person}{S{\'e}bastien Marcel}.} \bibinfo{year}{2018}\natexlab{}.
\newblock \showarticletitle{Deepfakes: a new threat to face recognition?
  assessment and detection}.
\newblock \bibinfo{journal}{\emph{arXiv preprint arXiv:1812.08685}}
  (\bibinfo{year}{2018}).
\newblock


\bibitem[Lee et~al\mbox{.}(2020)]%
        {9157722}
\bibfield{author}{\bibinfo{person}{Cheng-Han Lee}, \bibinfo{person}{Ziwei Liu},
  \bibinfo{person}{Lingyun Wu}, {and} \bibinfo{person}{Ping Luo}.}
  \bibinfo{year}{2020}\natexlab{}.
\newblock \showarticletitle{MaskGAN: Towards Diverse and Interactive Facial
  Image Manipulation}. In \bibinfo{booktitle}{\emph{2020 IEEE/CVF Conference on
  Computer Vision and Pattern Recognition (CVPR)}}.
  \bibinfo{pages}{5548--5557}.
\newblock
\urldef\tempurl%
\url{https://doi.org/10.1109/CVPR42600.2020.00559}
\showDOI{\tempurl}


\bibitem[Li et~al\mbox{.}(2020a)]%
        {9156865}
\bibfield{author}{\bibinfo{person}{Lingzhi Li}, \bibinfo{person}{Jianmin Bao},
  \bibinfo{person}{Hao Yang}, \bibinfo{person}{Dong Chen}, {and}
  \bibinfo{person}{Fang Wen}.} \bibinfo{year}{2020}\natexlab{a}.
\newblock \showarticletitle{Advancing High Fidelity Identity Swapping for
  Forgery Detection}. In \bibinfo{booktitle}{\emph{2020 IEEE/CVF Conference on
  Computer Vision and Pattern Recognition (CVPR)}}.
  \bibinfo{pages}{5073--5082}.
\newblock
\urldef\tempurl%
\url{https://doi.org/10.1109/CVPR42600.2020.00512}
\showDOI{\tempurl}


\bibitem[Li et~al\mbox{.}(2020b)]%
        {li2020celebdf}
\bibfield{author}{\bibinfo{person}{Yuezun Li}, \bibinfo{person}{Xin Yang},
  \bibinfo{person}{Pu Sun}, \bibinfo{person}{Honggang Qi}, {and}
  \bibinfo{person}{Siwei Lyu}.} \bibinfo{year}{2020}\natexlab{b}.
\newblock \showarticletitle{Celeb-df: A large-scale challenging dataset for
  deepfake forensics}. In \bibinfo{booktitle}{\emph{Proceedings of the IEEE/CVF
  Conference on Computer Vision and Pattern Recognition}}.
  \bibinfo{pages}{3207--3216}.
\newblock


\bibitem[Nirkin et~al\mbox{.}(2019)]%
        {9010341}
\bibfield{author}{\bibinfo{person}{Yuval Nirkin}, \bibinfo{person}{Yosi
  Keller}, {and} \bibinfo{person}{Tal Hassner}.}
  \bibinfo{year}{2019}\natexlab{}.
\newblock \showarticletitle{FSGAN: Subject Agnostic Face Swapping and
  Reenactment}. In \bibinfo{booktitle}{\emph{2019 IEEE/CVF International
  Conference on Computer Vision (ICCV)}}. \bibinfo{pages}{7183--7192}.
\newblock
\urldef\tempurl%
\url{https://doi.org/10.1109/ICCV.2019.00728}
\showDOI{\tempurl}


\bibitem[Petrov et~al\mbox{.}(2020)]%
        {Petrov2020DeepFaceLabAS}
\bibfield{author}{\bibinfo{person}{Ivan Petrov}, \bibinfo{person}{Daiheng Gao},
  \bibinfo{person}{Nikolay Chervoniy}, \bibinfo{person}{Kunlin Liu},
  \bibinfo{person}{Sugasa Marangonda}, \bibinfo{person}{Chris Um{\'e}},
  \bibinfo{person}{Mr. Dpfks}, \bibinfo{person}{RP Luis}, \bibinfo{person}{Jian
  Jiang}, \bibinfo{person}{Sheng Zhang}, \bibinfo{person}{Pingyu Wu},
  \bibinfo{person}{Bo Zhou}, {and} \bibinfo{person}{Weiming Zhang}.}
  \bibinfo{year}{2020}\natexlab{}.
\newblock \showarticletitle{DeepFaceLab: A simple, flexible and extensible face
  swapping framework}.
\newblock \bibinfo{journal}{\emph{ArXiv}}  \bibinfo{volume}{abs/2005.05535}
  (\bibinfo{year}{2020}).
\newblock


\bibitem[Rossler et~al\mbox{.}(2019)]%
        {rossler2019faceforensics}
\bibfield{author}{\bibinfo{person}{Andreas Rossler}, \bibinfo{person}{Davide
  Cozzolino}, \bibinfo{person}{Luisa Verdoliva}, \bibinfo{person}{Christian
  Riess}, \bibinfo{person}{Justus Thies}, {and} \bibinfo{person}{Matthias
  Nie{\ss}ner}.} \bibinfo{year}{2019}\natexlab{}.
\newblock \showarticletitle{Faceforensics++: Learning to detect manipulated
  facial images}. In \bibinfo{booktitle}{\emph{Proceedings of the IEEE/CVF
  International Conference on Computer Vision}}. \bibinfo{pages}{1--11}.
\newblock


\bibitem[Seferbekov(2020)]%
        {dfdc_solution}
\bibfield{author}{\bibinfo{person}{Selim Seferbekov}.}
  \bibinfo{year}{2020}\natexlab{}.
\newblock \bibinfo{booktitle}{\emph{DFDC 1st place solution}}.
\newblock
\urldef\tempurl%
\url{"https://github.com/selimsef/dfdc_deepfake_challenge"}
\showURL{%
\tempurl}


\bibitem[Siarohin et~al\mbox{.}(2019)]%
        {Siarohin2019FirstOM}
\bibfield{author}{\bibinfo{person}{Aliaksandr Siarohin},
  \bibinfo{person}{Stephane Lathuiliere}, \bibinfo{person}{S. Tulyakov},
  \bibinfo{person}{Elisa Ricci}, {and} \bibinfo{person}{N. Sebe}.}
  \bibinfo{year}{2019}\natexlab{}.
\newblock \showarticletitle{First Order Motion Model for Image Animation}.
\newblock \bibinfo{journal}{\emph{ArXiv}}  \bibinfo{volume}{abs/2003.00196}
  (\bibinfo{year}{2019}).
\newblock


\bibitem[Tan and Le(2021a)]%
        {tan2021efficientnetv2}
\bibfield{author}{\bibinfo{person}{Mingxing Tan} {and} \bibinfo{person}{Quoc~V.
  Le}.} \bibinfo{year}{2021}\natexlab{a}.
\newblock \bibinfo{title}{EfficientNetV2: Smaller Models and Faster Training}.
\newblock
\newblock
\showeprint[arxiv]{2104.00298}~[cs.CV]


\bibitem[Tan and Le(2021b)]%
        {https://doi.org/10.48550/arxiv.2104.00298}
\bibfield{author}{\bibinfo{person}{Mingxing Tan} {and} \bibinfo{person}{Quoc~V.
  Le}.} \bibinfo{year}{2021}\natexlab{b}.
\newblock \showarticletitle{EfficientNetV2: Smaller Models and Faster
  Training}.
\newblock  (\bibinfo{year}{2021}).
\newblock
\urldef\tempurl%
\url{https://doi.org/10.48550/ARXIV.2104.00298}
\showDOI{\tempurl}


\bibitem[Wodajo and Atnafu(2021)]%
        {wodajo2021deepfake}
\bibfield{author}{\bibinfo{person}{Deressa Wodajo} {and}
  \bibinfo{person}{Solomon Atnafu}.} \bibinfo{year}{2021}\natexlab{}.
\newblock \showarticletitle{Deepfake Video Detection Using Convolutional Vision
  Transformer}.
\newblock \bibinfo{journal}{\emph{arXiv preprint arXiv:2102.11126}}
  (\bibinfo{year}{2021}).
\newblock


\bibitem[Yang et~al\mbox{.}(2019)]%
        {yang2019exposing}
\bibfield{author}{\bibinfo{person}{Xin Yang}, \bibinfo{person}{Yuezun Li},
  {and} \bibinfo{person}{Siwei Lyu}.} \bibinfo{year}{2019}\natexlab{}.
\newblock \showarticletitle{Exposing deep fakes using inconsistent head poses}.
  In \bibinfo{booktitle}{\emph{ICASSP 2019-2019 IEEE International Conference
  on Acoustics, Speech and Signal Processing (ICASSP)}}. IEEE,
  \bibinfo{pages}{8261--8265}.
\newblock


\bibitem[Zhang et~al\mbox{.}(2016)]%
        {zhang2016joint}
\bibfield{author}{\bibinfo{person}{Kaipeng Zhang}, \bibinfo{person}{Zhanpeng
  Zhang}, \bibinfo{person}{Zhifeng Li}, {and} \bibinfo{person}{Yu Qiao}.}
  \bibinfo{year}{2016}\natexlab{}.
\newblock \showarticletitle{Joint face detection and alignment using multitask
  cascaded convolutional networks}.
\newblock \bibinfo{journal}{\emph{IEEE Signal Processing Letters}}
  \bibinfo{volume}{23}, \bibinfo{number}{10} (\bibinfo{year}{2016}),
  \bibinfo{pages}{1499--1503}.
\newblock


\end{thebibliography}

\appendix

\end{document}